\documentclass[10pt, a4paper]{article}

\usepackage{arabtex}
\usepackage{utf8}
\setcode{utf8}
\usepackage{graphicx}
\usepackage{subcaption}
\usepackage{multirow}
\usepackage{booktabs}
\usepackage{bm}
\usepackage{adjustbox}
\usepackage{enumitem}
\usepackage{graphicx}
\usepackage{longtable}
\usepackage{caption}

\usepackage{float} 

\usepackage{array}

\usepackage[final]{lrec2026} 

\title{Can LLMs Control Readability? \\ A Multi-Dimensional Evaluation Framework for CEFR-Controlled Arabic Generation }
\name{Nour Rabih, Chatrine Qwaider, Ted Briscoe} 

\address{Mohamed bin Zayed University of Artificial Intelligence (MBZUAI) \\
         \{nour.rabih, chatrine.qwaider, ted.briscoe\}@mbzuai.ac.ae\\}

\abstract{
While Large Language Models (LLMs) can generate fluent Arabic text, their ability to reliably control readability levels remains unclear. We propose a multi-dimensional evaluation framework for Common European Framework of Reference for Language (CEFR)-controlled Arabic text generation, assessing whether instruction-following LLMs can serve as reliable generators for adaptive language learning.
Our framework integrates controlled prompting, automatic readability prediction using a validated Taha-19 model, lexical constraint validation, and syntactic complexity profiling. Results show that structured prompting substantially improves CEFR alignment. In particular, CEFR-guided prompting with lexical constraints achieves the highest conformity to reference linguistic profiles (0.91 cosine similarity) and near-perfect agreement with predicted readability levels (0.99), while unconstrained prompting exhibits weak control.
These findings establish an empirical foundation for integrating readability-aware Arabic text generation into adaptive educational systems.
\\ \newline  }

\begin{document}

\maketitleabstract

\section{Introduction}
Large Language Models (LLMs) are increasingly used in educational technologies to generate reading materials, exercises, and language-learning content on demand. In such systems, controlling the linguistic difficulty of generated text is essential, as the Common European Framework of Reference for Languages (CEFR)~\cite{cefr2001}, which defines proficiency levels from A1 to C2, is widely adopted in educational settings. For safe deployment in learning environments, generative models must therefore produce not only fluent text, but content that is reliably aligned with learners’ reading abilities.
This challenge is particularly acute for Arabic. Its rich morphology, flexible word order, and syntactic variability make readability control substantially harder than in many high-resource languages. However, despite growing interest in readability-controlled generation \cite{ribeiro2023generating,trott2024measuring}, most existing work is based on prompting a LLM and relies on a single readability metric. This limits its applicability to adaptive learning systems, where generated content should be continuously assessed and adjusted in a fine-grained way to individual learners. 

In this work, we propose a multi-dimensional framework for CEFR-controlled Arabic text generation with LLMs, designed for integration into adaptive educational pipelines. Our framework combines controlled prompting, automatic readability prediction, CEFR-aligned lexical constraints, and syntactic complexity analysis to evaluate whether LLMs can generate Arabic texts that are fluent and pedagogically appropriate across proficiency levels.
We first establish a data-driven alignment between CEFR levels and the fine-grained 19-level \citet{Taha:2017:guidelines} readability scale by applying the pretrained BAREC readability model \cite{elmadani-etal-2025-large} to CEFR-labeled learner essays from the ZEABUC and ARWI Arabic Essay Scoring datasest \cite{habash2022zaebuc, qwaider2025enhancing}. In addition, we construct detailed linguistic profiles for each CEFR level based on these corpora, capturing characteristic syntactic and lexical properties. These empirically derived level-specific profiles are later used as evaluation benchmarks to systematically assess the structural and lexical alignment of generated essays with target proficiency levels.

We then prompt GPT-4o\footnote{\url{https://openai.com/index/hello-gpt-4o/}} to generate Arabic essays under different conditions, including prompts specifying only the target CEFR level and prompts augmented with CEFR-appropriate lexical and syntactic constraints. The outputs are evaluated using the BAREC model for fine-grained readability estimation and assessed against the constructed reference profiles to determine the extent to which their lexical and syntactic properties align with the expected specifications of each CEFR level.

Our contributions are threefold:
\begin{itemize}
    \item  The first multi-dimensional study of CEFR-controlled Arabic text generation with LLMs, framed within an adaptive learning perspective.
    \item  A principled alignment between CEFR and a fine-grained Arabic readability scale, enabling continuous modeling of text difficulty.
    \item  A dataset of CEFR-controlled Arabic essays\footnote{https://github.com/noorrabih/CEFR-Controlled-Arabic-Generation-Data.git} to support future research on personalized, readability-aware text generation for Arabic.
\end{itemize}

The rest of the paper is structured as follows. \S \ref{sec:related} reviews reviews
related work.  
\S \ref{sec:data} presents the  datasets used.
\S \ref{sec:methodology} describes the proposed evaluation framework,  
\S \ref{sec:experimental} details the experimental setup. 
\S \ref{sec:results} reports the results and analysis across prompting conditions. \S \ref{sec:conclusion}, concludes the paper and outlines directions for future work.

\section{Related Work}
\label{sec:related}

\paragraph{Arabic Readability}

Research on Arabic readability has explored a range of standards, datasets, and modeling approaches. Early work by \citet{khallaf2021automatic} modeled Arabic readability using a CEFR-inspired scheme with a coarse three-level classification. Other studies have adopted grade-based readability levels for first-language (L1) readers or instructional proficiency bands for second-language (L2) learners, reflecting curriculum-driven rather than standardized assessment criteria \cite{CAVALLISFORZA201838}. More recent dataset-driven efforts, such as DARES, provide both coarse-grained(4 levels) and fine-grained (12 levels) readability annotations, enabling more detailed modeling of Arabic text complexity \cite{el2024dares}. In parallel, the BAREC dataset \cite{elmadani-etal-2025-large} offers a fine-grained 19-level readability scale (Taha-19) designed specifically for Arabic, a pedagogically motivated framework inspired by \citet{Taha:2017:guidelines}, and is accompanied by a fine-tuned transformer-based model for automatic readability prediction.
Other transformer-based approaches have further advanced readability modeling, Readme++ introduces a multilingual CEFR classification model applicable to Arabic \cite{naous2024readme++}, however, its performance on Arabic is lower compared to higher-resource languages, highlighting the additional challenges associated with Arabic readability modeling. In this work, we leverage BAREC for fine-grained readability assessment, enabling evaluation of CEFR alignment and linguistic complexity in generated Arabic text.

\vspace{-7pt}
\paragraph{Controlled Text Generation and Readability Evaluation}

Prior work has investigated readability-controlled summarization and simplification, often using traditional readability formulas such as Flesch–Kincaid or Lexile scores \cite{ribeiro2023generating, trott2024measuring}. More recent studies have examined whether instruction-tuned LLMs can follow explicit readability constraints specified in prompts, evaluating generated text using trained SVMs from \citet{xia-etal-2016-text} to predict the CEFR level of the generated essay. 

Most existing analyses focus on English and rely on a single readability metric, limiting their ability to capture nuanced linguistic variation. Work on Arabic is especially scarce, and to our knowledge, no prior study has systematically evaluated CEFR-controlled Arabic text generation, nor examined syntactic complexity as an additional diagnostic signal.

\section{Data}
\label{sec:data}
We build our generation experiments and evaluation on two complementary data sources:
\vspace{-5pt}
\paragraph{ZAEBUC} The Zayed Arabic-English Bilingual Undergraduate Corpus (ZAEBUC)\footnote{https://sites.google.com/view/zaebuc/home} \cite{habash2022zaebuc}, is a publicly available dataset of short Arabic essays written by first-year university students and annotated for linguistic features including CEFR ratings. it provides 214 manually corrected Arabic essays with consistent CEFR labels, offering a reliable basis for studying proficiency.
\vspace{-5pt}
\paragraph{ARWI}\label{par:SED} The Arabic Read Write and Improve dataset\footnote{https://github.com/mbzuai-nlp/arabic-aes-bea25}\cite{qwaider2025enhancing}, is a publicly available dataset that contains synthetic generated essays for automated essay scoring .
  
 From this work, we use both the synthetic essays (3220 essays) and their associated leveled prompts and topics, which specify target proficiency levels. A few examples are shown in Table~\ref{tab:prompts}. In our experiments, these prompts and topics are reused to generate new essays using GPT-4o.
\begin{table}[t]
\centering
\footnotesize
\setlength{\tabcolsep}{2pt}
\begin{tabular}{p{1.5cm} p{3cm} p{3cm}}
\hline
\textbf{Level} & \textbf{Arabic Prompt} & \textbf{English Translation} \\
\hline
Beginner & \<اوصف يومك المفضل> & Describe your favorite day. \\
\hline
Interm. & \<كيف يمكننا التعامل\\ مع الضغوط النفسية؟> & How can we deal with psychological stress? \\
\hline
Advanced & \<تحدث عن أهمية التعليم \\الرقمي في عصرنا الحالي.> & Talk about the importance of digital education in our current era. \\
\hline
\end{tabular}
\caption{Example CEFR-aligned essay prompts}
\label{tab:prompts}
\end{table}
The original dataset organizes prompts into three broad proficiency categories: Beginner, Intermediate, and Advanced. To align with the CEFR framework, we map these categories into six CEFR levels as follows: Beginner → A1–A2, Intermediate → B1–B2, and Advanced → C1–C2. The dataset contains 161 prompts in total, distributed as 50 Beginner, 56 Intermediate, and 55 Advanced prompts.

We merge ARWI and ZAEBUC to create a unified proficiency-based essay collection. These datasets are selected because they are annotated and graded according to students’ writing proficiency levels, making them suitable for modeling learner ability. By combining them, we construct coherent proficiency profiles aligned with CEFR levels. Table~\ref{tab:bea_zaebuc_distribution} presents the distribution of essays after merging the datasets. The \textit{Unassessable} essays from ZAEBUC (6 instances) are excluded to ensure consistency across proficiency levels.

\begin{table}[h]
\centering
\begin{tabular}{lccc}
\hline
\textbf{CEFR Level} & \textbf{ARWI} & \textbf{ZAEBUC} & \textbf{Total} \\
\hline
A1 & 500 & 0   & 500 \\
A2 & 500 & 7   & 507 \\
B1 & 560 & 110 & 670 \\
B2 & 560 & 80  & 640 \\
C1 & 550 & 11  & 561 \\
C2 & 550 & 0   & 550 \\
Unassessable & 0 & 6 & 6 \\
\hline
\end{tabular}
\caption{Distribution of essays across CEFR levels for ARWI and ZAEBUC datasets.}
\label{tab:bea_zaebuc_distribution}
\end{table}

\vspace{-7pt}
\paragraph{SAMER Lexicon}
We also utilize the SAMER readability lexicon \citep{al-khalil-etal-2020-large}, a large-scale lexical resource for Modern Standard Arabic containing 26,578 manually annotated lemmas extracted from news and literary corpora. Each lemma is assigned one of five grade-based readability levels, annotated in triplicate by language professionals from different Arab regions. The lexicon combines corpus frequency information with expert judgment, providing a reliable lexical complexity signal that we incorporate into our proficiency modeling. To align SAMER with CEFR, we map its five levels to CEFR bands as follows: A1–A2 $\rightarrow$ Level 1, B1 $\rightarrow$ Level 2, B2 $\rightarrow$ Level 3, C1 $\rightarrow$ Level 4, and C2 $\rightarrow$ Level 5. This mapping preserves the ordinal progression of lexical complexity across both frameworks and enables consistent vocabulary constraints during controlled generation.\footnote{The mapping from SAMER levels to CEFR bands is heuristic and based on aligning the ordinal progression of lexical difficulty across the two frameworks. While this provides a practical approximation for controlled generation, more principled alignment strategies (e.g., data-driven calibration or joint modeling) remain an interesting direction for future work.}

\section{Methodology}
\label{sec:methodology}

\subsection{Assessment Standards}

We adopt two complementary readability standards to evaluate the generated texts: the \textbf{CEFR} and the fine-grained \textbf{Taha-19} readability scale. Together, these frameworks enable both coarse proficiency assessment and detailed analysis of linguistic complexity.


\paragraph{CEFR.}
The CEFR \cite{cefr2001} 
defines six proficiency levels (A1--C2) and is widely used to describe non-native learners’ linguistic attainment across reading, writing, listening, and speaking. It provides a standardized framework for assessing and comparing language proficiency across different educational contexts.

\paragraph{Taha-19.}
The Taha-19 scale is a fine-grained Arabic readability framework consisting of 19 levels, pedagogically motivated and derived from the Taha--Thomure model \cite{Taha:2017:guidelines}. It defines 19 ordered levels of reading difficulty tailored to Arabic, capturing fine-grained progression in linguistic complexity beyond coarse CEFR categories. Lower levels correspond to simple sentence structures, limited vocabulary, and minimal morphological variation, while higher levels reflect increased syntactic embedding, richer vocabulary, and more complex discourse structures. It is accompanied by the \textbf{BAREC} transformer-based predictor, which is trained on a large, balanced Arabic readability corpus. We use the BAREC-model to assign sentence-level Taha-19 readability scores.
\vspace{-7pt}

\paragraph{Taha-19 and CEFR Alignment.} To enable joint analysis across coarse CEFR levels and fine-grained readability, we establish an empirical alignment between the two scales using the ZAEBUC and ARWI corpora. Taha-19 readability scores are inferred using the BAREC model.\footnote{\url{https://huggingface.co/CAMeL-Lab/readability-arabertv2-d3tok-CE}}
Each essay is segmented into sentences using standard Arabic punctuation-based splitting (full stops, question marks, and exclamation marks). Sentence-level readability is inferred for each segment, and document-level readability is computed as the average of sentence-level predictions, yielding a continuous fine-grained score for every CEFR-labeled essay.

\begin{figure}
    \centering
    \includegraphics[width=1\linewidth]{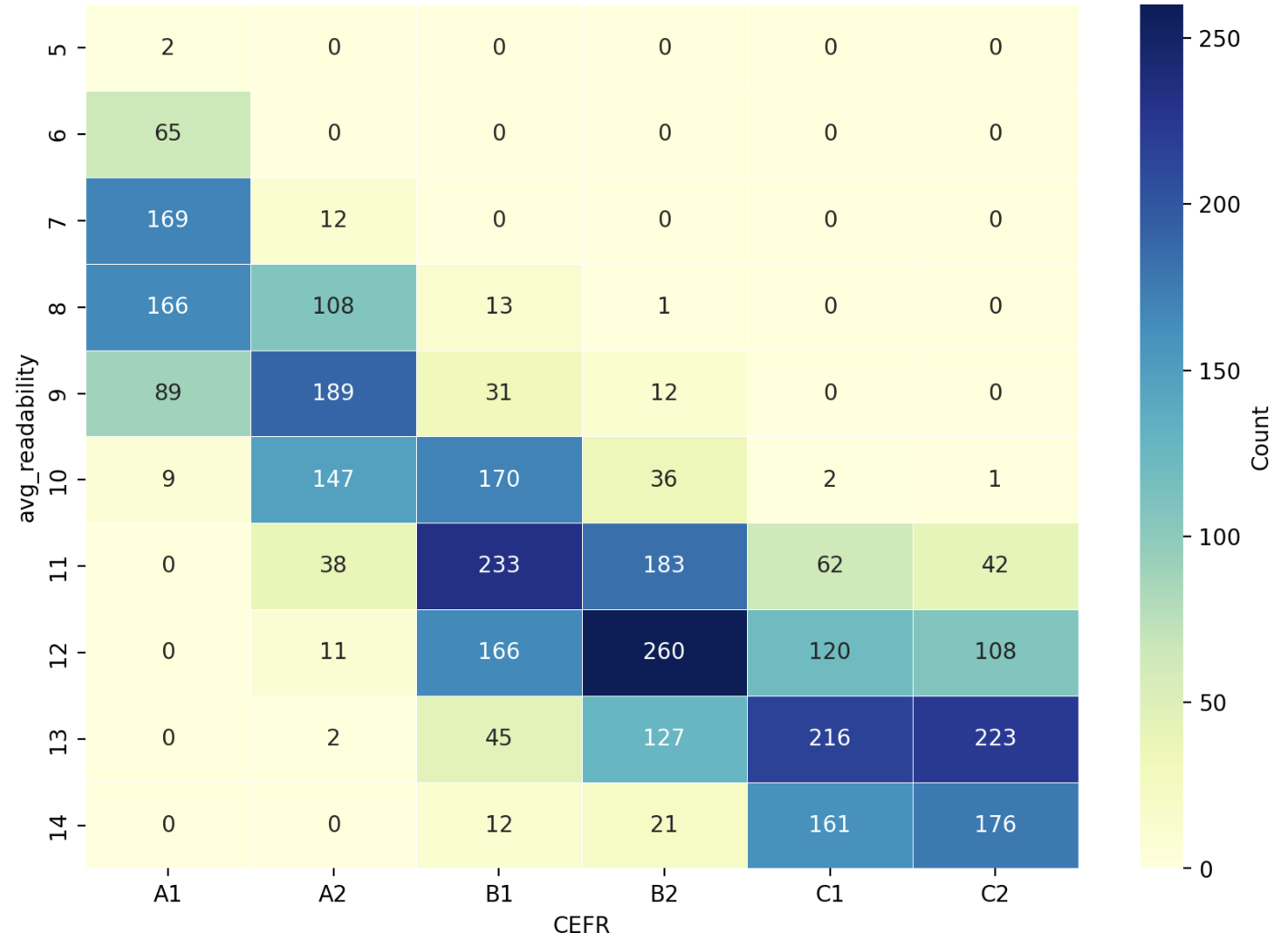}
    \caption{Taha-19 readability scores alignment with CEFR levels.}
    \label{fig:CEFR_BAREC}
\end{figure}

Figure \ref{fig:CEFR_BAREC} shows the distribution of Taha-19 readability levels across CEFR Labels. A clear trend is observed: A1–A2 texts exhibit the highest density in lower readability bands (ranging betweend 6–10), B1–B2 texts concentrate in intermediate ranges (ranging between 10–13), and C1–C2 texts align with higher bands (ranging between 12–14). This progression is further confirmed by the mean Taha-19 scores computed for each CEFR level: A1 (7.61) and A2 (9.26) correspond to lower readability bands, B1 (11.03) and B2 (11.81) fall within intermediate ranges, while C1 (12.78) and C2 (12.93) reflect higher readability levels. The steady increase in mean scores across CEFR levels demonstrates strong alignment between the Taha-19 scale and CEFR proficiency progression.
To statistically quantify this alignment, we compute the Spearman rank correlation between ordinal CEFR levels (A1–C2 mapped to 1–6) and document-level Taha-19 scores. The results show a strong positive correlation ($\rho$ = 0.84, $p$ < 0.001), indicating a significant monotonic relationship between CEFR proficiency progression and fine-grained readability.

\begin{figure*}
    \centering
    
    \begin{subfigure}{0.32\textwidth}
        \centering
        \includegraphics[width=\linewidth]{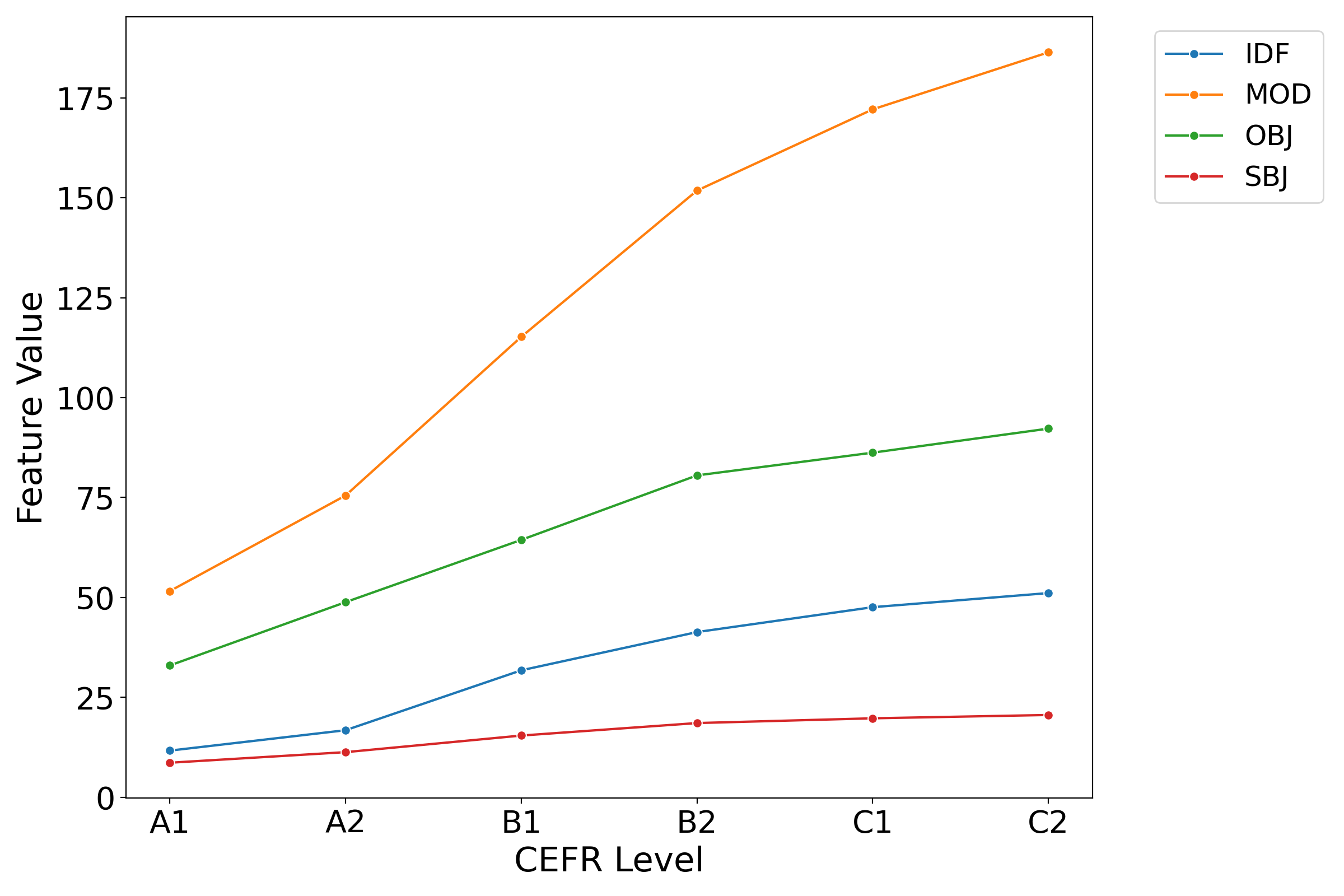}
        \caption{Dependency}
    \end{subfigure}
    \hfill
    \begin{subfigure}{0.32\textwidth}
        \centering
        \includegraphics[width=\linewidth]{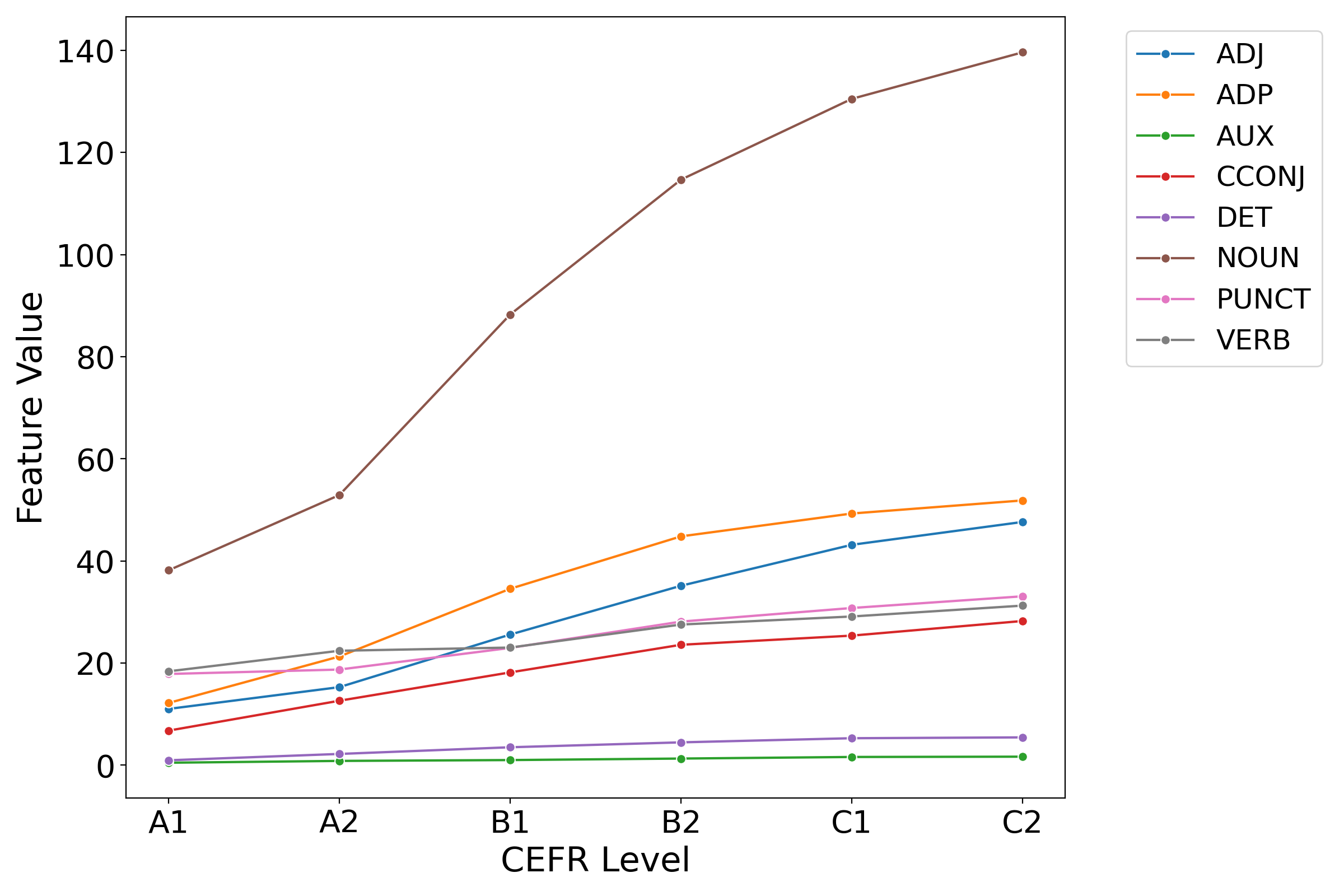}
        \caption{POS}
    \end{subfigure}
    \hfill
    \begin{subfigure}{0.32\textwidth}
        \centering
        \includegraphics[width=\linewidth]{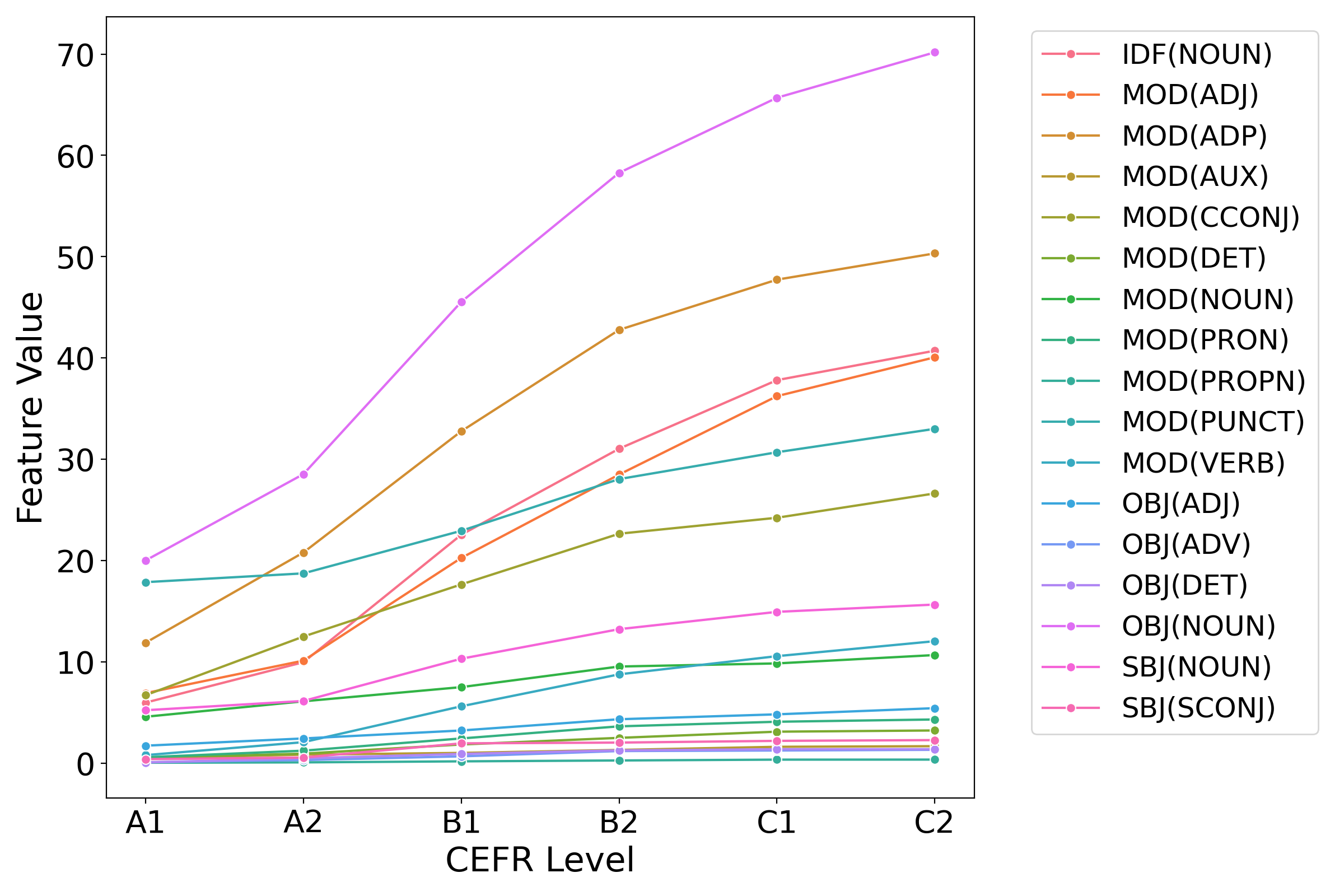}
        \caption{dep(POS)}
    \end{subfigure}

    \caption{Mean of Syntactic features across CEFR levels. (a) shows the progression of dependency relations, 
    (b) presents the distribution of POS tags, and (c) depicts the combined dependency--POS patterns.}
    \label{fig:syntactic_features}

\end{figure*}

\subsection{Essay Generation}

We formulate the generation task as producing Arabic essays on specified topics, enabling controlled evaluation of readability across CEFR levels. This setup reflects real-world usage, where educators and content creators rely on LLMs, such as GPT models, to generate pedagogical content aligned with learner proficiency levels \cite{kasneci2023chatgpt, whalen2023chatgpt}.

To examine whether LLMs can respond to increasingly explicit readability instructions, we design five instructional prompting styles that incrementally encode more information about the target reading level.

We consider five prompting conditions with increasing degrees of control. P1 provides no explicit readability information and serves as an unconstrained baseline. P2 specifies only the target CEFR level. P3 introduces syntactic control by imposing CEFR-aligned constraints on sentence structure and complexity based on the created profiles (Section \ref{subsec:Linguistic_profile}). P4 augments level specification with CEFR-aligned vocabulary constraints by requiring the inclusion of level-appropriate words.(Section~\ref{sec:vocab}).  Finally, P5 combines both vocabulary and syntactic constraints. The prompts used for each condition along with their translations are provided in Appendix~\ref{appendix:prompts}.

Due to the structure of the ARWI dataset prompts (Section~\ref{par:SED}), P1 is evaluated at three broad proficiency bands: A, B, and C. In contrast, P4 and P5 are evaluated at five finer-grained CEFR levels (A, B1, B2, C1, and C2), reflecting the granularity supported by the SAMER vocabulary classification system, which categorizes words into five proficiency levels.

\subsection{Linguistic profile Construction}
\label{subsec:Linguistic_profile}

\paragraph{Syntax}
To construct syntactic profiles for each CEFR level, we conduct an in-depth syntactic analysis on the data. Each essay is processed using the CamelParser\footnote{\url{https://github.com/CAMeL-Lab/camel_parser}, version 2.0} \cite{Elshabrawy:2023:camelparser}, from which we extract three categories of features: (1) Part-of-Speech (POS) tags, (2) dependency relations, and (3) combined dependency-POS patterns.

The analysis yields a total of 15 unique POS tags, 7 dependency relations, and 72 dependency–POS combinations. We then examine the distribution of these features across CEFR levels to identify meaningful and consistent trends, as illustrated in Figure~\ref{fig:syntactic_features}(a,b,c). 

Across these subfigures, features demonstrate clear monotonic or near-monotonic increases with proficiency level, indicating structural and syntactic development. Based on the consistency of these trends, we select 29 syntactic patterns that exhibit clear developmental progression and incorporate them into the level-specific syntactic profiles.
\vspace{-10pt}
\paragraph{Dependency Tree Depth} In addition to categorical syntactic features, we analyze structural complexity through dependency tree depth. For each sentence, we compute the depth of its dependency tree and then calculate the mean depth per essay. This aggregated measure serves as an indicator of syntactic embedding and hierarchical complexity across proficiency levels.

\begin{figure}
    \centering
    \includegraphics[width=1\linewidth]{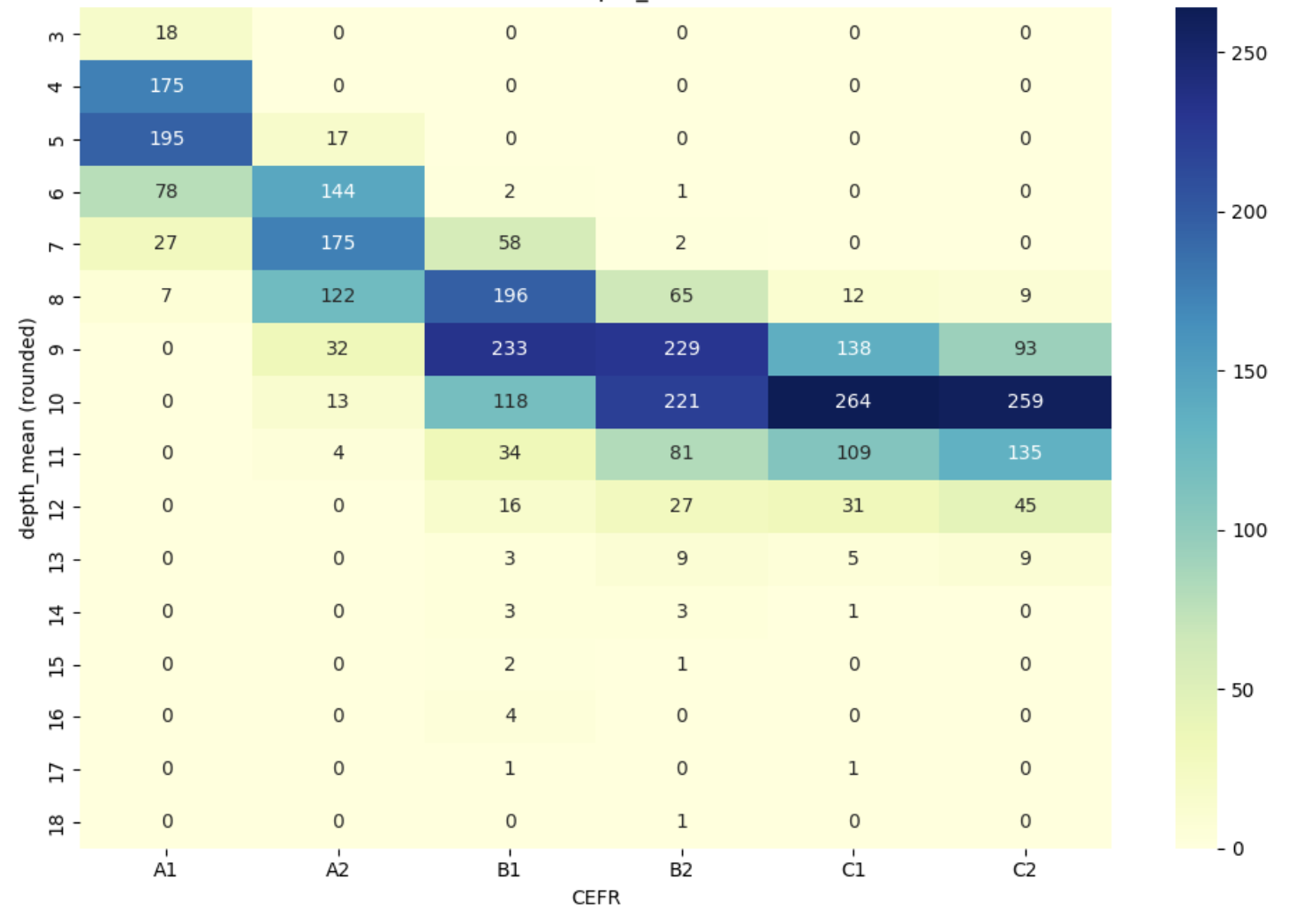}
    \caption{Distribution of mean dependency tree depths across CEFR levels}
    \label{fig:depth}
\end{figure}

The distribution of average essay dependency depths across CEFR levels is illustrated in Figure~\ref{fig:depth}. As shown in the figure, higher proficiency levels tend to exhibit greater tree depths, reflecting increased structural embedding and more complex hierarchical constructions. This trend is quantitatively confirmed by the mean dependency depth values for each CEFR level: A1 (4.90) and A2 (7.12) show comparatively shallow syntactic structures, B1 (8.97) and B2 (9.71) demonstrate moderate structural expansion, while C1 (10.05) and C2 (10.25) exhibit the deepest hierarchical constructions. The steady increase in mean depth across proficiency levels suggests that syntactic embedding grows progressively with CEFR advancement, supporting the role of dependency depth as an indicator of linguistic complexity.

\vspace{-5pt}
\paragraph{Lexical and Surface-Level}
\label{subsec:lexical_surface}

\begin{figure}
    \centering
    \includegraphics[width=1\linewidth]{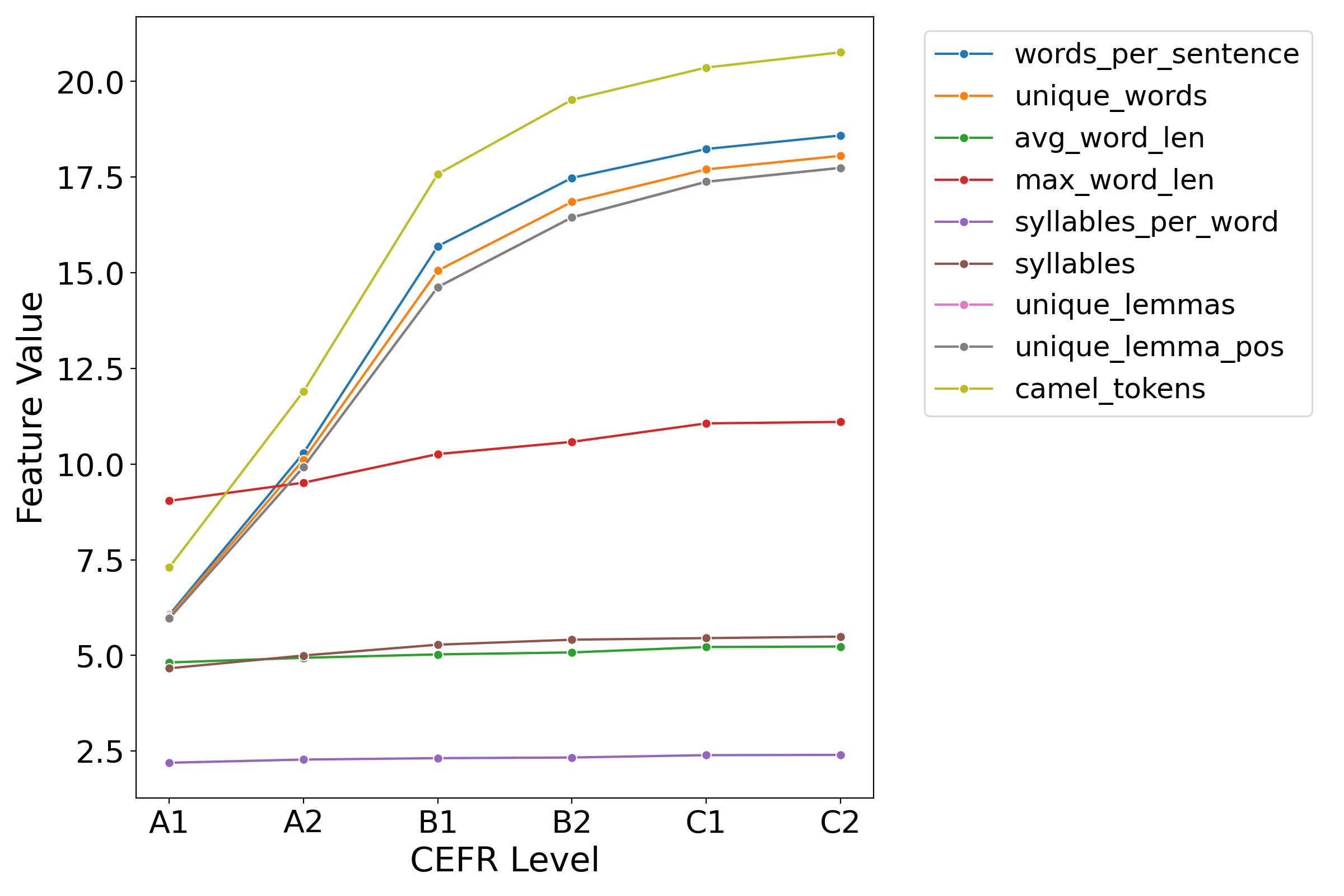}
    \caption{Lexical and surface-level features across CEFR levels.}
    \label{fig:lexical_surface_features}
\end{figure}

\vspace{-10pt}
In addition to syntactic profiling, we extract lexical and surface-level features to capture vocabulary richness, morphological variation, and overall textual complexity across CEFR levels. These features complement the syntactic rule set by providing quantitative indicators of lexical development. 

We compute core lexical statistics, including the total number of words, number of unique words, and average sentence length (words per sentence). To measure lexical diversity beyond surface forms, we calculate the number of unique lemmas and unique lemma–POS pairs, enabling analysis of conceptual vocabulary breadth and syntactic-functional variation.

We further model word-level complexity using length and syllable-based features, including average and maximum word length, as well as average and maximum syllable counts per word. These measures capture phonological and morphological complexity associated with proficiency progression. As shown in Figure~\ref{fig:lexical_surface_features}, these lexical and surface-level features exhibit a clear increasing trend across CEFR levels, reflecting systematic growth in lexical diversity, sentence length, and morphological complexity as proficiency advances from A1 to C2.

Together, these syntactic, structural, and lexical features define comprehensive proficiency-specific linguistic profiles, capturing both categorical and gradient aspects of language development. These profiles serve as reference benchmarks for evaluating generated texts and modeling CEFR-aligned complexity. A detailed summary of feature statistics across CEFR levels is provided in Appendix~\ref{app:profiles}.

\subsection{Vocabulary List Construction}
\label{sec:vocab}

To support lexical control in the constrained prompting condition, we automatically construct CEFR-aligned vocabulary lists for each topic. For a given topic and target CEFR level, we first prompt GPT to generate a candidate set of words using the template below:
\vspace{-5pt}
\begin{quote}
\small
You are an Arabic vocabulary instructor designing word lists aligned with CEFR levels.\\
Target CEFR level: \{cefr\}\\
Topic (Arabic): ``\{topic\}''\\
Task:
\begin{itemize}[noitemsep, topsep=0pt]
    \item Suggest a list of vocabulary suitable for a learner at level \{cefr\}, related to the topic above, that the learner can use to write an Arabic essay.
    \item Ensure the words are appropriate in frequency and difficulty for \{cefr\}.
    \item Output only the list of Arabic words, separated by commas.
    \item Number of words: between \{min\} and \{max\}.
\end{itemize}

Example format (no numbering, no extra text):\\
\texttt{word1, word2, word3, ...}
\end{quote}

The generated candidate lists are then filtered using \textbf{SAMER} \cite{al-khalil-etal-2020-large}, a five-level Arabic lexical difficulty lexicon.  Only words whose SAMER level matches the target CEFR band are retained.

If the filtered list contains fewer than the minimum required number of words (3 words), we repeat the generation-filtering process. If the list remains insufficient after multiple attempts, we apply a vocabulary relevance model \cite{reimers-2019-sentence-bert} that ranks SAMER words by semantic similarity to the topic and selects the highest-scoring items from the appropriate difficulty band. This fallback mechanism ensures that each topic is associated with a sufficient number of level-appropriate and semantically relevant lexical constraints.

A complete example of the constructed prompts and their corresponding CEFR-aligned vocabulary lists is provided in Appendix~\ref{appendix:prompts_vocab}.

\subsection{Evaluation Metrics}

\label{subsec:evaluation}

To evaluate how closely the generated readability-controlled texts match the reference CEFR-level profiles, we compute the cosine similarity between their feature vectors. Specifically, given two real-valued feature vectors $P_i$ (the CEFR reference profile) and $Q_i$ (the generated text), the cosine similarity is calculated as:
\begin{equation}
\cos(\theta) =
\frac{\sum_i P_i Q_i}
{\sqrt{\sum_i P_i^2} \cdot \sqrt{\sum_i Q_i^2}}
\end{equation}
Cosine similarity is particularly suitable in our setting because it measures the agreement between high-dimensional feature representations (over 40 linguistic features) in terms of their relative distributional patterns rather than absolute magnitudes. This is important for readability profiling, where we aim to capture how features co-vary across CEFR levels, rather than their raw values.
Compared to alternative measures such as Euclidean distance, which is sensitive to scale and absolute differences in individual features, cosine similarity focuses on the orientation of feature vectors in the shared space. This makes it more robust when combining heterogeneous linguistic features (e.g., counts, ratios, and averages) and better reflects whether the generated profiles follow the same structural trends as the reference CEFR profiles.

Before computing cosine similarity, we normalize all features using standard score normalization (z-score scaling) to ensure that features with larger numerical ranges do not dominate the similarity computation.
\vspace{-7pt}
\paragraph{Evaluation Granularity}

To provide a comprehensive analysis, we compute cosine similarity at multiple levels of granularity:

\begin{itemize}[noitemsep, topsep=0pt]
    \item \textbf{Overall}:  across all levels and features jointly.
    \item \textbf{ Per-cluster}: we group the profile features into 4 groups and evaluate each separately. Table~\ref{tab:feature_clusters} summarizes the feature groupings used in our analysis.
    \item \textbf{Per-level}:  across all features within each CEFR level.
    \item \textbf{Per-feature}: across CEFR levels for each individual feature.

\end{itemize}

\begin{table}[htbp]
\centering
\caption{Feature clusters used for evaluation}
\label{tab:feature_clusters}

\footnotesize
\setlength{\tabcolsep}{6pt}
\renewcommand{\arraystretch}{1.1}

\begin{tabular}{ll}
\toprule
\textbf{Cluster} & \textbf{Feature} \\
\midrule

\multirow{11}{*}{Surface}
& total words \\
& avg words per sentence \\
& total unique words \\
& avg unique words \\
& avg word length \\
& max word length \\
& avg syllables per word \\
& max syllables \\
& avg unique lemmas \\
& avg unique lemma pos \\
& avg camel tokens \\

\midrule

\multirow{4}{*}{Dependency}
& IDF \\
& MOD \\
& OBJ \\
& SBJ \\

\midrule

\multirow{8}{*}{POS}
& ADJ \\
& ADP \\
& AUX \\
& CCONJ \\
& DET \\
& NOUN \\
& PUNCT \\
& VERB \\

\midrule

\multirow{18}{*}{DepPOS}
& IDF(NOUN) \\
& MOD(ADJ) \\
& MOD(ADP) \\
& MOD(AUX) \\
& MOD(CCONJ) \\
& MOD(DET) \\
& MOD(NOUN) \\
& MOD(PRON) \\
& MOD(PROPN) \\
& MOD(PUNCT) \\
& MOD(VERB) \\
& OBJ(ADJ) \\
& OBJ(ADV) \\
& OBJ(DET) \\
& OBJ(NOUN) \\
& SBJ(NOUN) \\
& SBJ(SCONJ) \\
& Tree Depth \\

\bottomrule
\end{tabular}
\end{table}

\section{Experimental Setup}
\label{sec:experimental}
\subsection{Vocabulary List Construction}

\paragraph{Generation.}
For each prompt–CEFR pair, we generate candidate vocabulary lists using GPT-4o with temperature set to 0.6 to balance diversity and control. The number of requested words is CEFR-dependent: A1 (20–30), A2 (25–35), B1 (30–40), B2 (35–45), C1 (40–50), and C2 (40–60). The model is instructed to output comma-separated Arabic words only, without explanations or formatting.
\paragraph{SAMER Validation and Morphological Alignment.}
All generated words are morphologically analysed using the CAMeL Tools \footnote{\url{https://github.com/CAMeL-Lab/camel_tools}, version 1.2.0} \cite{obeid-etal-2020-camel} to obtain lemma–POS representations. These are matched against the SAMER readability lexicon to verify that each word belongs to the intended difficulty band. Words whose SAMER readability does not match the target level are discarded. For multi-word expressions, difficulty is determined by the maximum SAMER level among their components. If the filtered list contains fewer than three valid words, the generation–validation cycle is repeated to ensure sufficient lexical constraints.
\vspace{-15pt}
\paragraph{Vocabulary Relavance Fallback}
In cases where repeated validation still yields insufficient vocabulary items, we apply a semantic relevance model based on Sentence-BERT (sentence-transformers/paraphrase-multilingual-MiniLM-L12-v2) \cite{reimers-2019-sentence-bert}. The model computes cosine similarity between the prompt topic and candidate SAMER lemmas within the appropriate readability level. Words exceeding a similarity threshold 0.50 and meeting minimum frequency constraints (frequency information is part of the SAMER Lexicon) are selected. These items are merged with previously validated words to produce the final vocabulary list used in constrained prompting.

\begin{table*}
\centering
\caption{Cosine Similarity across dimensions.}
\label{tab:cosine_results}

{
\begin{tabular}{lllccccc}
\toprule
\textbf{Category} & \textbf{Group} & \textbf{Dimension / Level} & \textbf{P1} & \textbf{P2} & \textbf{P3} & \textbf{P4} & \textbf{P5} \\
\midrule

\multirow{1}{*}{Overall}
& -- & Overall
& 0.10 & 0.74 & \textbf{0.91} & 0.65 & 0.66 \\

\midrule

\multirow{4}{*}{Cluster}
& -- & Surface
& 0.21 & 0.76 & \textbf{0.91} & 0.83 & 0.86 \\
& -- & Dependency
& 0.11 & 0.77 & \textbf{0.98} & 0.63 & 0.89 \\
& -- & POS
& -0.01 & 0.69 & \textbf{0.90} & 0.48 & 0.62 \\
& -- & DepPOS
& 0.08 & 0.73 & \textbf{0.90} & 0.61 & 0.56 \\

\midrule

\multirow{6}{*}{Per Level}
& \multirow{2}{*}{A} & A1
& \multirow{2}{*}{-0.77} 
& 0.87 
& \textbf{0.95} 
& \multirow{2}{*}{0.69} 
& \multirow{2}{*}{0.79} \\
&  & A2
&  
& 0.36 
& \textbf{0.90} 
&  
&  \\

& \multirow{2}{*}{B} & B1
& \multirow{2}{*}{0.76}
& -0.22 
& -0.23 
& -0.56 
& \textbf{0.59} \\
&  & B2
&  
& \textbf{0.96} 
& 0.84 
& 0.89 
& 0.37 \\

& \multirow{2}{*}{C} & C1
& \multirow{2}{*}{0.97}
& \textbf{0.97} 
& 0.90 
& 0.90 
& 0.58 \\
&  & C2
&  
& \textbf{0.97} 
& 0.96 
& 0.89 
& 0.78 \\
\bottomrule
\end{tabular}
}
\end{table*}
\vspace{-7pt}
\subsection{Generation}
All essays are generated using GPT-4o through the Chat Completions API in batch mode. Each batch entry corresponds to a single topic–CEFR pair. Each request includes a fixed system instruction in Arabic  \<(أنت مساعد يكتب مقالات إنشائية عربية بمستويات \\ قرائية مختلفة.)>, \textit{(You are an assistant that writes Arabic essays at different readability levels.)} and the prompt. Generation is performed with a temperature of 0.7 to allow moderate linguistic variation while preserving control over readability constraints. All other parameters remain at default settings, and each prompt is generated once per prompting condition.

\subsection{Profile Construction}

The Dependency, POS and DepPOS cluster features listed in Table \ref{tab:feature_clusters} are extracted using the CAMeL dependency parser \cite{Elshabrawy:2023:camelparser}. These features are computed through straightforward counting and averaging at the CEFR level, as described in Section \ref{sec:methodology}. Similarly, surface-level length features are obtained using simple counts, such as word counts and length-based measures.
To extract morphological and lexical features, we use the CAMeL Tools \cite{obeid-etal-2020-camel} Modern Standard Arabic (MSA) BERT-based morphological disambiguation pipeline. 
This pipeline provides rich linguistic annotations, including diacritized forms, lemmas, part-of-speech tags, and morphological attributes. Lemma-based features are directly extracted from the disambiguator outputs. Tokenization is performed using CAMeL’s \texttt{simple\_word\_tokenize} function to obtain the count of tokens.
Syllable counts are computed by incorporating both phonological and morphological information. Specifically, we rely on the CAPHI (Consonant–Vowel pattern) representation (extracted from the disambiguation), the diacritized surface form, and morphological prefix annotations to obtain accurate syllable estimates. 
The CAPHI representation is segmented and examined for vowel units, with each vowel corresponding to a potential syllable. To improve accuracy, we apply the following linguistically motivated rules, following the approach of \citet{rabih2025noor}:
\begin{itemize}
\item Final vowels corresponding to inflectional diacritics (\<حركات الإعراب>) are excluded, as they reflect grammatical inflection rather than intrinsic word structure.
\item Morphological prefixes such as the definite article (\<ال التعريف>) and coordinating conjunction (\<واو العطف>) are excluded, since they function as clitics and do not contribute to the core syllabic structure of the lexical stem.

\item When CAPHI information is unavailable, syllables are estimated using a fallback heuristic that counts vowel-indicating diacritic characters in the diacritized word form.

\end{itemize}

\section{Results and Analysis}
\label{sec:results}

We evaluate all five prompting conditions (P1–P5) using automatic readability prediction (BAREC), and feature-level deviation analysis against the reference CEFR-aligned profiles.

Table \ref{tab:cosine_results} reports cosine similarity between generated texts under the five prompting conditions (P1–P5) and the CEFR reference feature profiles. For robustness, evaluation is performed at the level of aggregated profiles rather than individual essays. Specifically, for each prompting condition and CEFR level, we compute the average feature vector across all generated essays for that level, and cosine similarity is then calculated between this averaged generated profile and the corresponding CEFR reference profile.

Overall, P3 achieves the highest alignment (0.91) and consistently outperforms other prompting strategies across feature clusters, particularly in the Dependency cluster (0.98), indicating strong syntactic conformity. At beginner levels (A1–A2), P3 shows the strongest alignment, while P2 performs competitively at advanced levels (C1–C2).
Across feature categories, surface features show the most stable and consistently high alignment across prompting conditions, reflecting their direct responsiveness to explicit instructions. In contrast, syntactic features exhibit greater variability, while fine-grained DepPOS patterns remain the most challenging to control, as they emerge implicitly from writing style rather than being directly specifiable in prompts.
However, the model appears to struggle at Level B1, where cosine similarities are comparatively low and even negative for several prompting conditions. This behavior can be explained by a systematic shift in the generated profiles toward higher complexity: instead of matching the reference B1 distribution, the model tends to produce texts with feature values closer to B2. As a result, many features that are below the reference mean at B1 become above the mean in the generated outputs, leading to opposite directional deviations after normalization and consequently negative cosine similarity. This indicates that the model overestimates intermediate-level complexity, effectively collapsing B1 toward the adjacent higher proficiency level.

Notably, prompting strategies that include explicit vocabulary additions tend to reduce overall performance compared to structure-focused prompting. This indicates that simply increasing lexical content does not necessarily improve CEFR alignment and may disrupt feature balance. One possible explanation is that under structural prompting, the model can generate text more naturally, loosely adhering to syntactic guidelines while maintaining fluent sentence construction. In contrast, when required to incorporate specific lexical items, the model tends to construct sentences around these words rather than generating text that organically conforms to the target syntactic profile. This effect is reflected in the sharp drop in POS cluster alignment under P4 (0.48) compared to P3 (0.90), as well as the decrease in DepPOS alignment from 0.90 to 0.61.

\begin{table}[t]
\centering
\renewcommand{\arraystretch}{0.8}
\caption{Average Taha-19 level cosine similarity per prompt.}
\resizebox{\columnwidth}{!}{
\label{tab:barec_avg_cosine}

\begin{tabular}{lccccc}
\toprule
\textbf{Dimension} & \textbf{P1} & \textbf{P2} & \textbf{P3} & \textbf{P4} & \textbf{P5} \\
\midrule
Taha-19 level & 0.26 & 0.83 & \textbf{0.99} & 0.93 & 0.91 \\
\bottomrule
\end{tabular}}
\end{table}

Table~\ref{tab:barec_avg_cosine} reports the average cosine similarity between predicted BAREC levels and the reference CEFR levels across prompting conditions. Consistent with the feature-level analysis, P3 achieves the highest alignment (0.995), indicating near-perfect agreement with the target readability levels. P4 and P5 also show strong performance, while P2 performs moderately well. In contrast, P1 exhibits substantially lower alignment (0.2611). The large performance gap between P1 and P2 highlights the importance of explicit readability conditioning in LLM prompting. In P1, the model receives no signal about the target proficiency level, and therefore defaults to generating text at an average or internally preferred complexity level, which does not align with CEFR-specific linguistic profiles. These results further support the effectiveness of structured prompting in improving CEFR-level conformity.

A detailed table reporting the cosine similarity for each feature independently across prompting conditions is included in Appendix \ref{app:eval}.

\section{Conclusions and Future Work}
\label{sec:conclusion}
In this work, we presented the first multi-dimensional evaluation of CEFR-controlled Arabic text generation using LLMs. We proposed a comprehensive framework that combines automatic readability prediction (BAREC), lexical constraint validation, syntactic profiling, and feature-level cosine similarity analysis to assess whether instruction-following models can reliably generate CEFR-aligned Arabic texts.
Our findings show that structured prompting substantially improves readability control. In particular, syntactically guided prompting (P3) consistently achieved the highest alignment with CEFR reference profiles across overall, cluster-level, and fine-grained BAREC evaluations.
This study establishes an empirical foundation for integrating readability-aware generation into adaptive Arabic learning systems. By combining coarse CEFR categorization with fine-grained Taha-19 modeling, our framework enables scalable and interpretable evaluation of controlled text generation.

A promising direction for future research is extending this framework to the personalized learner level. Rather than relying solely on static CEFR-level profiles, individual student data could be used to construct learner-specific linguistic profiles, capturing strengths and weaknesses in vocabulary, syntax, and structural complexity. Generation prompts could then be dynamically adapted based on these personalized profiles, enabling fine-grained readability control tailored to individual learners. Such an approach is contingent upon the availability of sufficient and ethically collected learner data.
Additionally, integrating readability-aware generation into real adaptive tutoring systems would enable longitudinal assessment of learning impact. In a “read–write–improve” loop, learners could receive automatically generated texts aligned to their evolving proficiency, produce written responses, and obtain feedback informed by the same profiling framework. Over time, this would allow continuous monitoring of linguistic development and systematic adjustment of generated materials. Evaluating such systems in real educational environments would provide deeper insight into how readability-controlled generation affects learning outcomes.
Another direction for future work is evaluating multiple large language models. Since this study uses only GPT-4o, extending the analysis to other models would help assess the generalizability of the framework and determine whether the observed effects of structured prompting are consistent across different LLMs.
\vspace{-10pt}
\section*{Ethics Statement}

This work focuses on improving the reliability and safety of Arabic readability-controlled text generation for educational applications. All training and evaluation data used in this study are drawn from publicly available corpora (ZAEBUC, ARWI, SAMER, and BAREC), and no personally identifiable information was collected or processed.

\section*{Limitations}
Despite providing the first multi-dimensional evaluation of CEFR-controlled Arabic text generation, this study has several limitations. First, although BAREC shows strong alignment with CEFR progression, our evaluation relies on a single readability model; future work could incorporate additional readability predictors to further validate and compare results.

Second, experiments are conducted using a single LLM (GPT-4o), which limits the generalizability of our findings. Future research should evaluate multiple LLMs to assess robustness across different architectures and decoding strategies.

\section{Bibliographical References}\label{sec:reference}

\bibliographystyle{lrec2026-natbib}
\bibliography{lrec2026-example}

\appendix
\onecolumn

\section{CEFR Linguistic Profiles}
\label{app:profiles}

\begin{table}[htbp]
\centering
\caption{ CEFR Linguistic Profiles}
\label{tab:cefr_full_features}

\resizebox{\textwidth}{!}{
\begin{tabular}{lcccccc}
\toprule
\textbf{Feature} & \textbf{A1} & \textbf{A2} & \textbf{B1} & \textbf{B2} & \textbf{C1} & \textbf{C2} \\
\midrule

total\_words & 85.35 & 118.03 & 179.87 & 232.50 & 262.46 & 281.57 \\
avg\_words\_per\_sentence & 6.06 & 10.29 & 15.69 & 17.48 & 18.23 & 18.59 \\
total\_unique\_words & 84.73 & 116.19 & 174.47 & 225.18 & 255.03 & 273.60 \\
avg\_unique\_words & 6.02 & 10.12 & 15.06 & 16.85 & 17.70 & 18.06 \\
overall\_avg\_word\_len & 4.82 & 4.94 & 5.03 & 5.08 & 5.22 & 5.23 \\
overall\_max\_word\_len & 9.04 & 9.51 & 10.26 & 10.58 & 11.06 & 11.10 \\
avg\_syllables\_per\_word & 2.20 & 2.30 & 2.34 & 2.35 & 2.41 & 2.42 \\
max\_max\_syllables & 4.67 & 5.02 & 5.29 & 5.42 & 5.45 & 5.49 \\
avg\_unique\_lemmas & 5.97 & 9.93 & 14.62 & 16.44 & 17.37 & 17.74 \\
avg\_unique\_lemma\_pos & 5.97 & 9.93 & 14.63 & 16.44 & 17.38 & 17.74 \\
avg\_camel\_tokens & 7.31 & 11.90 & 17.58 & 19.52 & 20.36 & 20.76 \\

dep\_IDF\_mean & 11.69 & 16.79 & 31.78 & 41.34 & 47.57 & 51.10 \\
dep\_MOD\_mean & 51.60 & 75.51 & 115.20 & 151.79 & 172.11 & 186.37 \\
dep\_OBJ\_mean & 32.99 & 48.82 & 64.41 & 80.54 & 86.22 & 92.24 \\
dep\_SBJ\_mean & 8.66 & 11.30 & 15.46 & 18.58 & 19.77 & 20.60 \\

pos\_ADJ\_mean & 11.03 & 15.29 & 25.59 & 35.15 & 43.17 & 47.64 \\
pos\_ADP\_mean & 12.19 & 21.29 & 34.56 & 44.82 & 49.29 & 51.86 \\
pos\_AUX\_mean & 0.49 & 0.85 & 1.01 & 1.31 & 1.62 & 1.68 \\
pos\_CCONJ\_mean & 6.78 & 12.63 & 18.18 & 23.59 & 25.38 & 28.25 \\
pos\_DET\_mean & 0.97 & 2.21 & 3.52 & 4.48 & 5.29 & 5.45 \\
pos\_NOUN\_mean & 38.17 & 52.92 & 88.26 & 114.68 & 130.50 & 139.61 \\
pos\_PUNCT\_mean & 17.87 & 18.74 & 22.98 & 28.13 & 30.79 & 33.08 \\
pos\_VERB\_mean & 18.39 & 22.42 & 23.04 & 27.55 & 29.13 & 31.26 \\

depPOS\_IDF(NOUN)\_mean & 5.98 & 9.97 & 22.55 & 31.04 & 37.81 & 40.72 \\
depPOS\_MOD(ADJ)\_mean & 6.92 & 10.12 & 20.26 & 28.49 & 36.23 & 40.07 \\
depPOS\_MOD(ADP)\_mean & 11.89 & 20.80 & 32.77 & 42.78 & 47.73 & 50.33 \\
depPOS\_MOD(AUX)\_mean & 0.49 & 0.85 & 1.01 & 1.31 & 1.62 & 1.68 \\
depPOS\_MOD(CCONJ)\_mean & 6.71 & 12.50 & 17.65 & 22.65 & 24.23 & 26.63 \\
depPOS\_MOD(DET)\_mean & 0.53 & 0.94 & 1.84 & 2.50 & 3.11 & 3.23 \\
depPOS\_MOD(NOUN)\_mean & 4.58 & 6.11 & 7.51 & 9.54 & 9.85 & 10.67 \\
depPOS\_MOD(PRON)\_mean & 0.59 & 1.23 & 2.44 & 3.64 & 4.09 & 4.31 \\
depPOS\_MOD(PROPN)\_mean & 0.04 & 0.08 & 0.18 & 0.27 & 0.35 & 0.35 \\
depPOS\_MOD(PUNCT)\_mean & 17.87 & 18.74 & 22.94 & 28.05 & 30.70 & 33.01 \\
depPOS\_MOD(VERB)\_mean & 0.79 & 2.07 & 5.62 & 8.77 & 10.57 & 12.04 \\

depPOS\_OBJ(ADJ)\_mean & 1.73 & 2.44 & 3.23 & 4.34 & 4.82 & 5.43 \\
depPOS\_OBJ(ADV)\_mean & 0.09 & 0.31 & 0.67 & 1.20 & 1.25 & 1.32 \\
depPOS\_OBJ(DET)\_mean & 0.07 & 0.48 & 0.91 & 1.26 & 1.37 & 1.38 \\
depPOS\_OBJ(NOUN)\_mean & 20.01 & 28.54 & 45.54 & 58.28 & 65.69 & 70.17 \\

depPOS\_SBJ(NOUN)\_mean & 5.23 & 6.14 & 10.31 & 13.23 & 14.94 & 15.66 \\
depPOS\_SBJ(SCONJ)\_mean & 0.40 & 0.56 & 1.96 & 2.03 & 2.20 & 2.28 \\

tree\_depth\_mean & 4.90 & 7.12 & 8.97 & 9.71 & 10.05 & 10.25 \\
avg\_ Taha-19\_level & 7.61 & 9.26 & 11.03 & 11.81 & 12.78 & 12.93 \\

\bottomrule
\end{tabular}
}
\end{table}

\newpage
\section{Feature-Level Evaluation}
\label{app:eval}
\vspace{-20pt}

\setlength{\tabcolsep}{10pt}      
\renewcommand{\arraystretch}{1.25} 
\begin{center}
\normalsize 
\begin{longtable}{>{\raggedright\arraybackslash}p{6.2cm}ccccc}
\caption{Feature-Level Cosine Similarity per Prompt}
\label{tab:appendix_feature_cosine} \\

\toprule
\textbf{Feature} & \textbf{P1} & \textbf{P2} & \textbf{P3} & \textbf{P4} & \textbf{P5} \\
\midrule
\endfirsthead

\toprule
\textbf{Feature} & \textbf{P1} & \textbf{P2} & \textbf{P3} & \textbf{P4} & \textbf{P5} \\
\midrule
\endhead

\bottomrule
\endfoot

avg\_camel\_tokens & 0.07 & 0.71 & 0.98 & 0.89 & 0.92 \\
avg\_level & 0.26 & 0.83 & 1.00 & 0.93 & 0.91 \\
avg\_syllables\_per\_word & 0.85 & 0.85 & 0.62 & 0.99 & 0.83 \\
avg\_unique\_lemma\_pos & 0.08 & 0.71 & 0.99 & 0.88 & 0.91 \\
avg\_unique\_lemmas & 0.08 & 0.71 & 0.99 & 0.88 & 0.91 \\
avg\_unique\_words & 0.08 & 0.72 & 0.98 & 0.89 & 0.92 \\
avg\_words\_per\_sentence & 0.08 & 0.72 & 0.98 & 0.89 & 0.93 \\
dep\_IDF\_mean & 0.17 & 0.81 & 0.99 & 0.73 & 0.98 \\
dep\_MOD\_mean & 0.05 & 0.77 & 0.99 & 0.70 & 0.89 \\
dep\_OBJ\_mean & -0.02 & 0.63 & 0.98 & 0.46 & 0.91 \\
dep\_SBJ\_mean & 0.36 & 0.90 & 0.99 & 0.67 & 0.81 \\
depPOS\_IDF(NOUN)\_mean & 0.35 & 0.88 & 0.99 & 0.81 & 0.98 \\
depPOS\_MOD(ADJ)\_mean & 0.41 & 0.91 & 0.96 & 0.91 & 0.92 \\
depPOS\_MOD(ADP)\_mean & 0.06 & 0.74 & 0.99 & 0.63 & 0.94 \\
depPOS\_MOD(AUX)\_mean & -0.13 & 0.60 & 0.51 & -0.20 & 0.05 \\
depPOS\_MOD(CCONJ)\_mean & 0.03 & 0.72 & 0.94 & 0.74 & 0.47 \\
depPOS\_MOD(DET)\_mean & -0.03 & 0.76 & 0.94 & 0.49 & 0.75 \\
depPOS\_MOD(NOUN)\_mean & 0.05 & 0.60 & 0.99 & 0.45 & 0.88 \\
depPOS\_MOD(PRON)\_mean & -0.33 & 0.60 & 0.95 & 0.75 & 0.18 \\
depPOS\_MOD(PROPN)\_mean & -0.28 & 0.87 & 0.96 & 0.66 & 0.63 \\
depPOS\_MOD(PUNCT)\_mean & -0.09 & 0.70 & 0.88 & 0.45 & 0.39 \\
depPOS\_MOD(VERB)\_mean & -0.07 & 0.68 & 0.97 & 0.67 & 0.63 \\
depPOS\_OBJ(ADJ)\_mean & 0.48 & 0.87 & 0.69 & 0.79 & 0.75 \\
depPOS\_OBJ(ADV)\_mean & 0.13 & 0.52 & 0.88 & 0.64 & 0.52 \\
depPOS\_OBJ(DET)\_mean & 0.93 & 0.83 & 0.94 & 0.92 & 0.59 \\
depPOS\_OBJ(NOUN)\_mean & 0.06 & 0.76 & 0.99 & 0.63 & 0.93 \\
depPOS\_SBJ(NOUN)\_mean & 0.47 & 0.94 & 0.95 & 0.78 & 0.82 \\
depPOS\_SBJ(SCONJ)\_mean & 1.00 & 0.95 & 0.95 & 0.92 & 0.65 \\
depth\_mean & -0.02 & 0.65 & 0.97 & 0.84 & 0.94 \\
max\_max\_syllables & 0.00 & 0.62 & 0.90 & 0.49 & 0.62 \\
overall\_avg\_word\_len & 1.00 & 0.95 & 0.62 & 0.95 & 0.99 \\
overall\_max\_word\_len & 0.53 & 0.92 & 0.99 & 0.94 & 0.97 \\
pos\_ADJ\_mean & 0.40 & 0.91 & 0.95 & 0.88 & 0.96 \\
pos\_ADP\_mean & 0.07 & 0.75 & 0.99 & 0.64 & 0.96 \\
pos\_AUX\_mean & -0.13 & 0.60 & 0.51 & -0.20 & 0.05 \\
pos\_CCONJ\_mean & 0.01 & 0.72 & 0.96 & 0.72 & 0.67 \\
pos\_DET\_mean & 0.14 & 0.76 & 0.96 & 0.68 & 0.97 \\
pos\_NOUN\_mean & 0.16 & 0.82 & 0.99 & 0.69 & 0.94 \\
pos\_PUNCT\_mean & -0.08 & 0.70 & 0.88 & 0.46 & 0.40 \\
pos\_VERB\_mean & -0.24 & 0.40 & 0.94 & 0.15 & 0.20 \\
total\_unique\_words & 0.11 & 0.79 & 0.99 & 0.66 & 0.92 \\
total\_words & 0.11 & 0.79 & 0.99 & 0.67 & 0.93 \\

\end{longtable}
\end{center}

\section{CEFR Prompts and Vocabulary Lists}
\label{appendix:prompts_vocab}

The data below can be useful for various educational applications \footnote{https://github.com/noorrabih/CEFR-Controlled-Arabic-Generation-Data.git}.
 
\begin{figure*}[ht]
    \centering
\includegraphics [width=0.99\linewidth]
{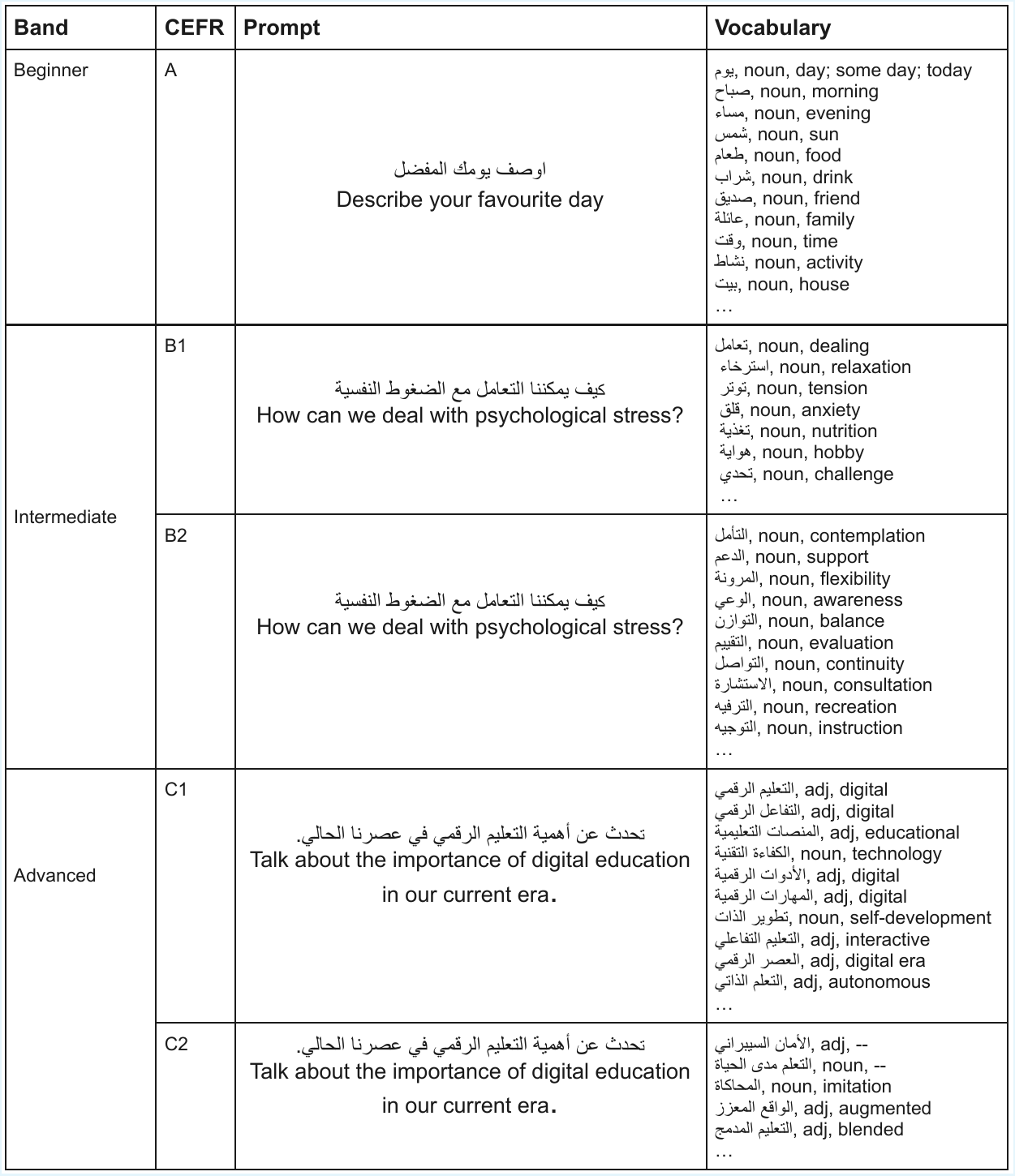}
\caption{Example of CEFR-aligned essay prompts with vocabulary lists.}

 \end{figure*}  

\setlength{\tabcolsep}{6pt}
\renewcommand{\arraystretch}{1.1}

\newpage
\section{Prompts}
\label{appendix:prompts}

\begin{figure*}[ht]
    \centering

\includegraphics [width = 0.99 \linewidth]
{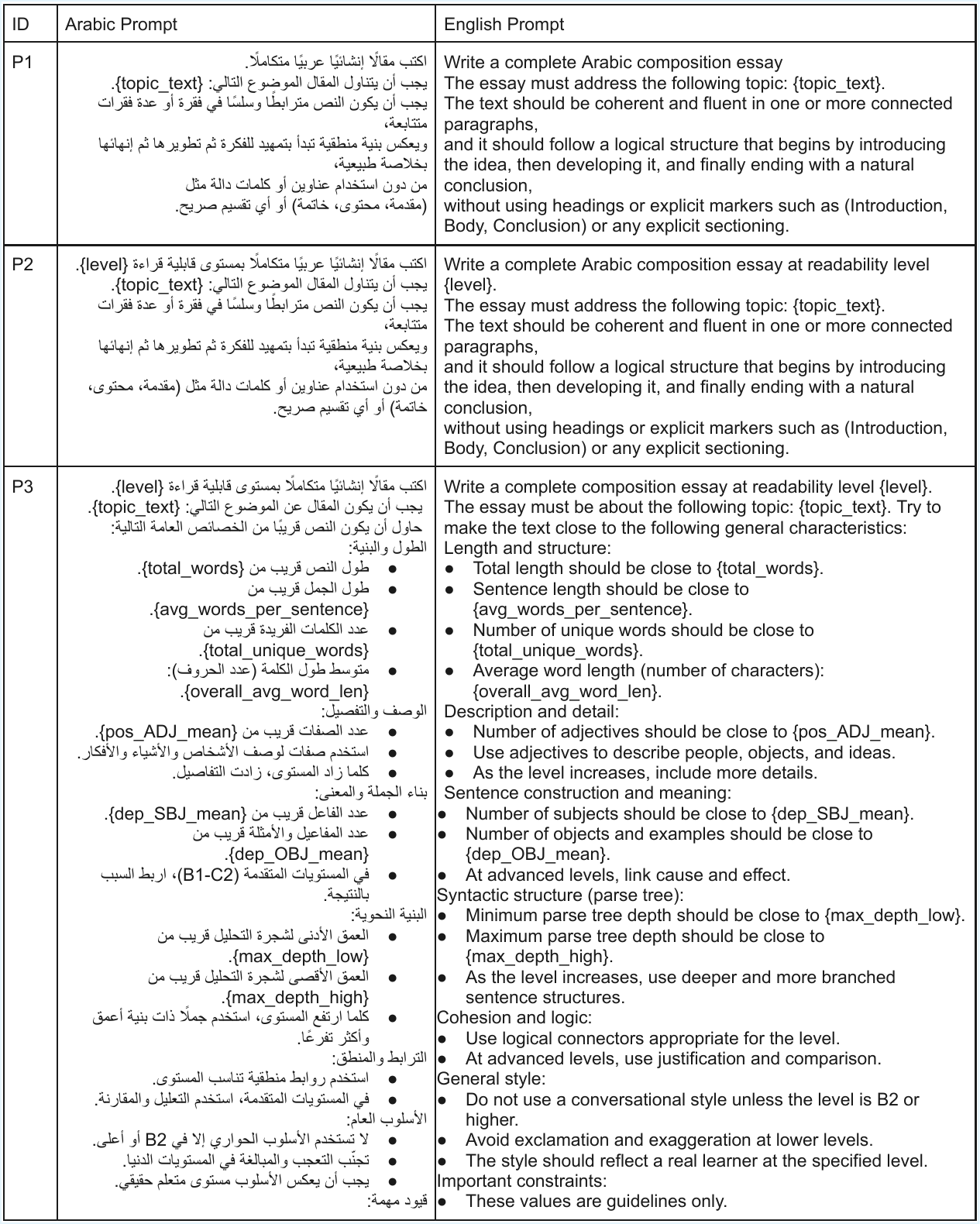}

 \end{figure*}  
\begin{figure*}[ht]
    \centering
\includegraphics [width=0.99\linewidth]
{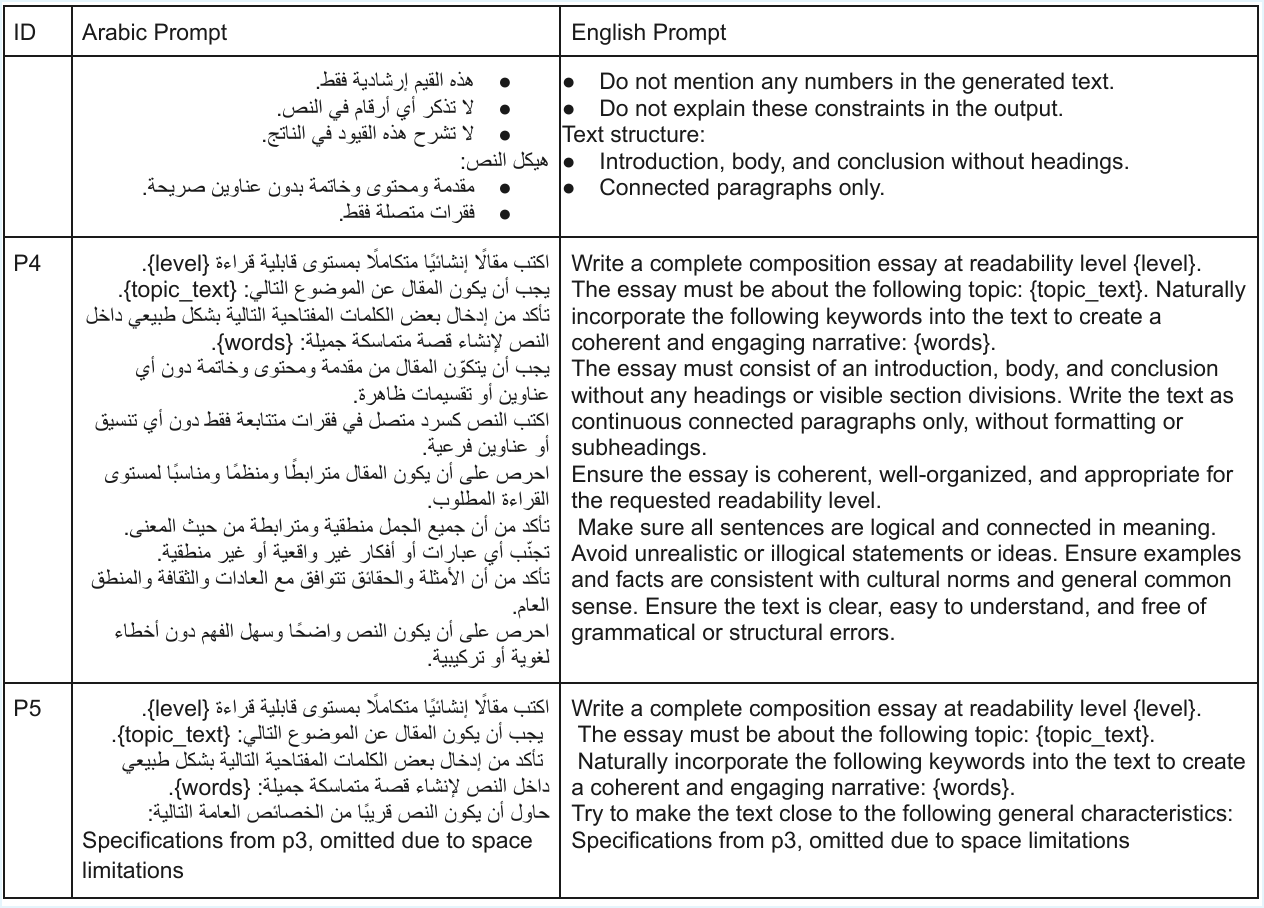}
\caption{Prompts used for controlled Arabic text generation at different readability levels.}

 \end{figure*}  

\setlength{\tabcolsep}{8pt}
\renewcommand{\arraystretch}{1.1}

\newcolumntype{L}[1]{>{\raggedright\arraybackslash}p{#1}}
\renewcommand{\arraystretch}{1.15}

\end{document}